\documentclass{ecai}
\usepackage{graphicx}
\usepackage{latexsym}
\usepackage[noadjust]{cite}
\usepackage{amsmath,amssymb,amsfonts}
\usepackage{algorithmic}
\usepackage{graphicx}
\usepackage{textcomp}
\usepackage{xcolor}
\usepackage[subtle]{savetrees}
\usepackage{multirow}
\usepackage{url}
\usepackage{pgfplots}
\pgfplotsset{width=6.5cm, height=8cm}
\def\BibTeX{{\rm B\kern-.05em{\sc i\kern-.025em b}\kern-.08em
    T\kern-.1667em\lower.7ex\hbox{E}\kern-.125emX}}

\newcommand{\model}{\textsc{dsgrl}}
\newcommand{\dgcnn}{\textsc{DGCNN}}
\newcommand{\diffpool}{\textsc{DiffPool}}

\newcommand{\ecc}{\textsc{ECC}}
\newcommand{\gin}{\textsc{GIN}}
\newcommand{\sage}{\textsc{GraphSage}}

\newcommand{\graphcl}{\textsc{GraphCL}}
\newcommand{\infograph}{\textsc{InfoGraph}}
\newcommand{\adgcl}{\textsc{adgcl}}

\newcommand{\dgi}{\textsc{DGI}}
\newcommand{\mvgrl}{\textsc{MVGRL}}
\newcommand{\gca}{\textsc{GCA}}

\newcommand{\byol}{\textsc{BGRL}}

\newcommand{\gcn}{\textsc{gcn}}
\newcommand{\gat}{\textsc{gat}}
\newcommand{\cluster}{\textsc{ClusterGCN}}
\newcommand{\saint}{\textsc{GraphSaint}}
\newcommand{\pprgo}{\textsc{pprgo}}
\newcommand{\oor}{\textsc{oor}}

\newcommand{\nodetovec}{\textsc{node2vec}}
\newcommand{\esim}{\textsc{ESim}}
\newcommand{\metapathtovec}{\textsc{metapath2vec}}
\newcommand{\herec}{\textsc{HERec}}
\newcommand{\han}{\textsc{han}}
\newcommand{\magnn}{\textsc{magnn}}

\begin{document}

\begin{frontmatter}

\title{Data-Driven Self-Supervised Graph Representation Learning} 

\author[A]{\fnms{Ahmed}~\snm{E. Samy}\thanks{Corresponding Author. Email: aesy@kth.se}}
\author[A]{\fnms{Zekarias}~\snm{T. Kefato}}
\author[A]{\fnms{\v{S}ar\={u}nas}~\snm{Girdzijauskas}} 

\address[A]{KTH, Royal Institute of Technology, Stockholm, Sweden}
\address{aesy@kth.se, zekarias@kth.se, sarunasg@kth.se}

\begin{abstract}
Self-supervised graph representation learning (SSGRL) is a representation learning paradigm used to reduce or avoid
manual labeling.
An essential part of SSGRL is graph data augmentation.
Existing methods usually rely on heuristics commonly identified through trial and error and are effective only within some application domains.
Also, it is not clear why one heuristic is better than another.
Moreover, recent studies have argued against some techniques (e.g., \emph{dropout}: that can change the properties of molecular graphs
or destroy relevant signals for graph-based document classification tasks).

In this study, we propose a novel data-driven SSGRL approach that automatically learns a suitable graph augmentation from the signal encoded in the graph (i.e., the nodes' predictive feature and topological information).
We propose two complementary approaches that produce learnable feature and topological augmentations.
The former learns multi-view augmentation of node features, and the latter learns a high-order view of the topology.
Moreover, the augmentations are jointly learned with the representation. Our approach is general that it can be applied to homogeneous and heterogeneous graphs.
We perform extensive experiments on node classification (using nine homogeneous and heterogeneous datasets)  and graph property prediction (using another eight datasets).
The results show that the proposed method matches or outperforms the SOTA SSGRL baselines and performs similarly to semi-supervised methods. 
The anonymised source code is available at \url{https://github.com/AhmedESamy/dsgrl/}\end{abstract}

\end{frontmatter}

\section{Introduction}
\label{sec:intro}
\begin{figure*}[ht!]
    \centering
    \includegraphics[scale=0.29, angle =270 ]{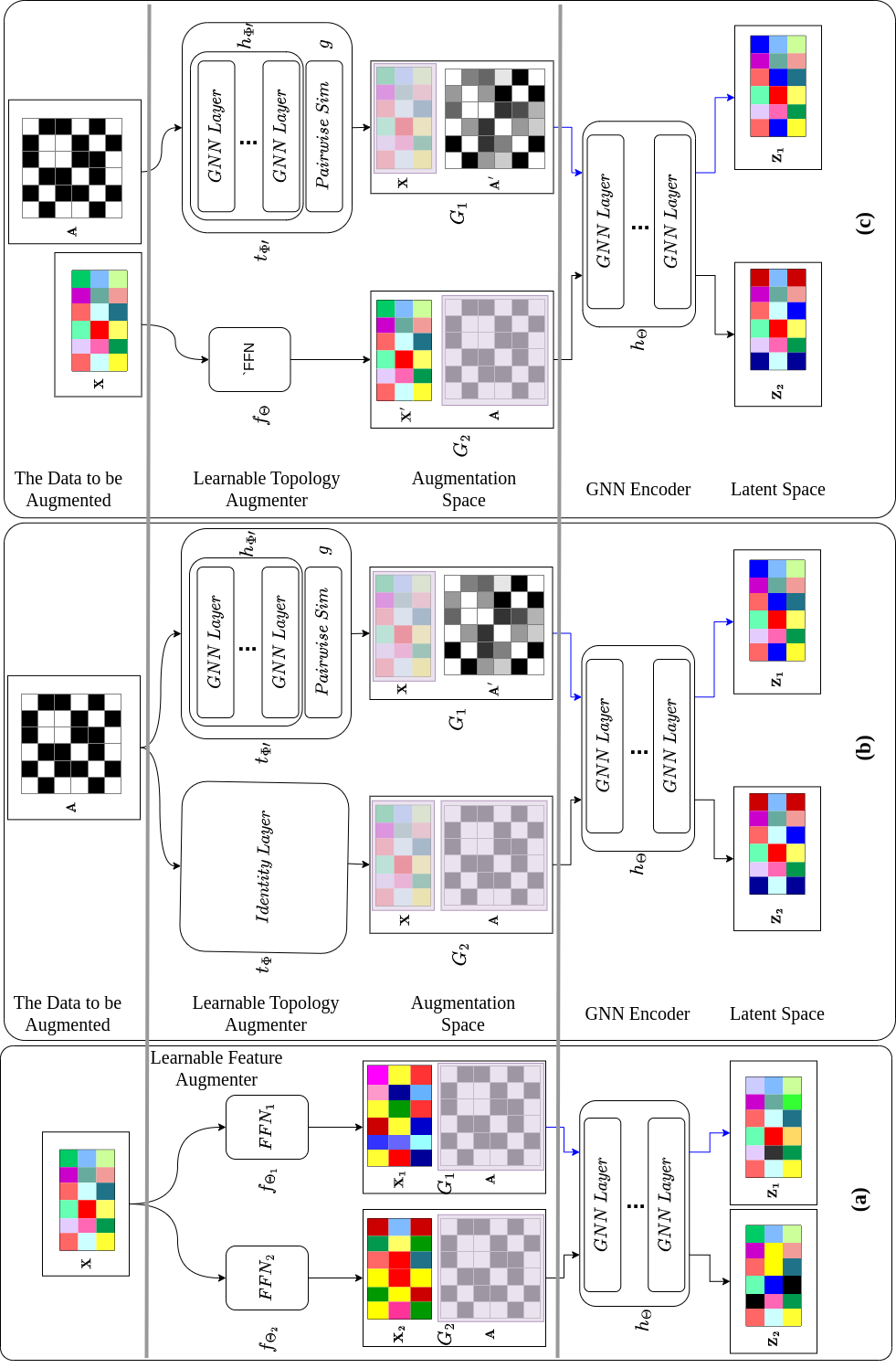}
    \caption{The Architecture of~\model~. The left figure (a) is based on learnable feature augmentation, where we only augment node features, $\mathbf{X}$. Two learnable augmenters $f_{\Theta_1}(\mathbf{X})$ and $f_{\Theta_2}(\mathbf{X})$ are applied on $\mathbf{X}$ to obtain the learned augmentations $\mathbf{X}_1$ and $\mathbf{X}_2$, respectively. The middle figure (b) is used for learning topological augmentation. Two augmentations, an identical, $t_\Phi(\mathbf{A})$, and learned ones, $t_{\Phi'}(\mathbf{A})$, are respectively applied on the adjacency matrix to obtain $\mathbf{A}$ and a high-order network $\mathbf{A}'$. The right figure (c) is used for combining learnable feature and topological augmentations. The augmentation results in two views $G_1 = (\mathbf{A}_1, \mathbf{X}_1)$ and $G_2 = (\mathbf{A}_2, \mathbf{X}_2)$ from the augmentation space, where $\mathbf{A}_1, \mathbf{X}_1, \mathbf{A}_2, \mathbf{X}_2$ are set to different values as shown in the above figures based on the augmentation type. Finally, a shared GNN encoder $h_\Theta$ is applied on the views, $h_\Theta(\mathbf{A}_1, \mathbf{X}_1)$ and $h_\Theta(\mathbf{A}_2, \mathbf{X}_2)$, respectively, to obtain the latent representations $\mathbf{Z}_1$ and $\mathbf{Z}_2$.
    The data modality ($\mathbf{A}$ or $\mathbf{X}$) that is not affected by an augmenter is blurred.}
    \label{fig:architecture}
\end{figure*}
Self-supervised graph representation learning (SSGRL) has been successfully used for graph representation learning (GRL)~\cite{10.1145/2623330.2623732,grover2016node2vec,10.1145/2736277.2741093,kefato2020gossip, samy2022schemawalk} in various domains where labeled data is scarce and manual label is expensive.
It has recently attracted interest across domains by achieving a competitive performance when compared to
semi-supervised approaches.
Considering the scarcity of labeled data, SSGRL has emerged as a new
paradigm that narrows down the performance gap between the unsupervised and semi-supervised learning methods.

Self-supervised learning (SSL) is commonly formulated as a \emph{predictive} or \emph{contrastive} 
learning~\cite{xie2021selfsupervised}.
For predictive models~\cite{devlin2019bert}, one first defines a related task on which an SSL model is pre-trained to extract meaningful patterns. The pre-trained model is subsequently refined  (fine-tuned) on a relevant but specific task of interest. Typically, an SSL model is pre-trained over large data as a starting point.
The quintessential models, particularly from NLP, are the ones that are pre-trained on masked word prediction tasks and are fine-tuned on other relevant tasks, such as text classification or translation.

On the other hand, contrastive models learn based on augmented views of a data point (e.g., image, graph) that are 
generated by applying a \emph{meaningful} perturbation on the original data point.
The representation of a data point is then learned by maximizing the mutual information between latent representations obtained from its augmented views.
The main challenge here is to produce augmented views of the data points.

The key to learning high-quality representations based on augmentation is that the perturbations should preserve 
semantics~\cite{DBLP:journals/corr/abs-2103-03230,bardes2021vicreg,thakoor2021bootstrapped,DBLP:journals/corr/abs-2011-10566}. 
For instance, a perturbation applied to an image of a dog should preserve \emph{``dogness"}.
Effective augmentation techniques for images (e.g., rotation, flipping, resizing) allow learning high-quality 
visual representations because they preserve the semantics of the original image. 
This is also true for SSL techniques in NLP~\cite{devlin2019bert,liu2019roberta,lan2020albert}, (e.g. synonym augmentation and word masking), 
such techniques do not alter the meaning of the original sentence.

Due to the complex nature of graph data, it is much more challenging to find appropriate techniques for augmenting graphs.
While some techniques are proposed, there is no standard technique that works well for graphs in different domains ~\cite{trivedi2021augmentations,10.1145/3442381.3449802,suresh2021adversarial,DBLP:journals/corr/abs-2010-13902,thakoor2021bootstrapped,velickovic2018deep,DBLP:journals/corr/abs-2006-05582,DBLP:journals/corr/abs-2103-14958}. 
Consequently, most efforts rely on finding a heuristic by trial and error to identify a suitable augmentation
for the graph at hand.

Generally, there are two classes of perturbations, which either corrupt the topology of the graph or 
node features.
The topology can be corrupted by dropping nodes and edges or adding new edges either randomly or through a
diffusion process~\cite{velickovic2018deep,10.1145/3442381.3449802,suresh2021adversarial,DBLP:journals/corr/abs-2006-05582}.
Similarly, dropout, masking, and permutation techniques have been used for corrupting node features~\cite{suresh2021adversarial,DBLP:journals/corr/abs-2103-14958}.
Nonetheless, it is unclear why a particular augmentation technique works better. A study~\cite{trivedi2021augmentations} has shown that these strategies are susceptible to destroying task-relevant information. 
Furthermore, in some cases, e.g., for molecular graphs, dropout techniques alter the semantics of the 
graph~\cite{xie2021selfsupervised}.
The effectiveness of such techniques usually comes not from the graph augmentations but from the strong inductive bias of the underlying learning algorithm,
particularly Graph Neural Networks (GNNs)~\cite{trivedi2021augmentations}.

In this paper, we follow a data-driven approach, where the augmentation process is guided by the inherent 
signal encoded in the graph.
Such an approach establishes obvious benefits, first as one can avoid trial and error in identifying a suitable augmentation
mechanism.
Second, it provides a flexible framework that can be adapted to different domains. 

Thus, we propose a novel \textbf{D}ata-driven \textbf{S}elf-supervised \textbf{G}raph \textbf{R}epresentation 
\textbf{L}earning (\model) method.
\model~is data-driven because it jointly learns the augmentation with the representation.
Similar to existing methods, we aim to augment either the topology or node features; however, unlike them~\model~learns both augmentations from the data. \model is a \textit{general} approach that can be applied to both homogeneous and heterogeneous graphs (i.e., graphs containing 
multiple node/edge types). 

Generally, for a given graph $G$, and a family of augmentation heuristics $\mathcal{A}$, existing methods apply either a 
topological, $A_t \sim \mathcal{A}$, or feature, $A_f \sim \mathcal{A}$, augmentation sampled from $\mathcal{A}$.
However,~\model~does not rely on $\mathcal{A}$, instead it learns $A_{t_\Phi}$ and $A_{f_\Theta}$; where $\Phi$ and
$\Theta$ are the learnable parameters of the topological and feature augmenters, respectively.
We materialize $t_\Phi$ using a GNN to obtain high-order node features; based on similarity scores between these features, we
compute a high-order weighted network as a topological augmentation.
The feature augmentation, $f_\Theta$, is simply a feed-forward neural network (FFN).
Note that, $A_{f_\Theta}$ in Fig. \ref{fig:architecture}(a) and $A_{t_\Phi}$  in Fig. \ref{fig:architecture}(b) are technically complementary and \emph{not competitive}. One can combine both augmentations at the same time as shown in Fig. \ref{fig:architecture}(c).
Next, given two augmented views $G_1$ and $G_2$ of a graph $G$, the latent representation $\mathbf{Z} = AGG(\mathbf{Z}_1, \mathbf{Z_2})$
of the graph is obtained by applying a shared GNN -- $h_\theta$, $\mathbf{Z}_1 = h_\theta(G_1)$ and $\mathbf{Z}_2 = h_\theta(G_2)$. 
$G_1$ and $G_2$ are learned using either topological or feature augmentation or a combination thereof.

Contrastive models require a negative (contrastive) term to prevent collapse, and negative sampling is the most 
common strategy for this~\cite{10.1145/3442381.3449802,zhu2020deep,DBLP:journals/corr/abs-2006-09882,DBLP:journals/corr/abs-1911-05722,wang2021selfsupervised,sun2021mocl,suresh2021adversarial,DBLP:journals/corr/abs-2106-07594,DBLP:journals/corr/abs-2006-05582,sun2020infograph,velickovic2018deep}.
However, it has two limitations; first, it requires a large batch size, and second, sampling truly negative 
terms is difficult.
Alternatives that do not require explicit negative sampling have been proposed to overcome such limitations~\cite{DBLP:journals/corr/abs-2011-10566,thakoor2021bootstrapped,DBLP:journals/corr/abs-2103-14958}.
Nevertheless, the alternatives are usually based on engineering tricks.
In this study, we closely follow a recent method that uses a principled approach based on variance, invariance and covariance
to prevent collapse~\cite{bardes2021vicreg}.

We perform extensive experiments using nine node classification datasets, including homogeneous and heterogeneous datasets, and another eight graph property predictions. 
We compare our method against seven popular SOTA SSGRL methods.
The results show that~\model~matches or improves the SSGRL baselines, and it is comparable with the semi-supervised methods. 


\section{The Proposed Method}
\label{sec:method}

\subsection{Preliminary}
We consider a graph ${G = (\mathbf{A}, \mathbf{X})}$ with a set of $N$ nodes $V$ and $M$ edges $E$.
$\mathbf{A} \in \lbrace 0, 1 \rbrace^{N \times N}$ 
denotes the adjacency matrix of $G$ and $\mathbf{X} \in \mathbb{R}^{N \times F}$ is a feature matrix, where
$F$ is the number of features.
For a given row index $i$, $\mathbf{A}[i] = \mathbf{a}_i$ and $ \mathbf{X}[i] = \mathbf{x}_i$ represent the 
topological structure and feature signals of node $i$.
For any index $i$, $\mathbf{A}[:,i] = \mathbf{a}_{:i}$ and $\mathbf{X}[:,i] = \mathbf{x}_{:i}$ refers to the 
indexing of the $i^{th}$ column of the adjacency and feature matrices, respectively.
Finally, $\mathbf{z}_{ij}$ corresponds to the $ij^{th}$ entry of any matrix $\mathbf{Z}$.

We consider a message passing GNN, $h_\Theta$, is given, and
\[
h_\Theta(G) = h_\Theta(\mathbf{A}, \mathbf{X}) = \sigma (\ldots \sigma(\mathbf{A}'\mathbf{X}^{(l)}\mathbf{W}^{(l)}) \ldots)
\]
where, $\mathbf{W}^{(l)} \in \Theta$ is the weight matrix of the $l^{th}$ layer, $\sigma$ is an activation function, e.g., \texttt{ReLU}, and
$\mathbf{A}'$ is a transformed adjacency matrix.
Depending on the type of GNN, one can apply different transformations on $\mathbf{A}$, e.g., the symmetric normalization used in~\cite{kipf2017semisupervised}.

\subsection{The Case for \model}
Several techniques for self-supervised graph representation learning (SSGRL) rely on perturbations by
randomly dropping nodes, edges, or subgraphs.
This perturbation is acceptable for social graphs. However, they are susceptible to
losing semantics.
For instance, dropping a node (an atom) or an edge (a bond) from a particular molecule could alter the 
essential properties of the molecule~\cite{xie2021selfsupervised}.
Furthermore, a recent study~\cite{trivedi2021augmentations} shows that even for other tasks, e.g., document representation learning based on graphs, 
dropout techniques could destroy task-relevant information.

We propose graph data augmentation techniques that are inspired by recent studies that advocate for learning
augmentation governed by the graph signal or context.

\subsection{Learnable Augmentation}
\label{sub:sec:learnable:augmentation}
The key hypothesis behind learning augmentations is that because it is a data-driven approach, it 
enables us to effectively capture augmentation signals without the human intervention needed for heuristics
based on trial and error.
Therefore, we propose two alternative approaches, which are learnable \emph{feature} and 
\emph{topology} augmentation.

\subsubsection{Learnable Feature Augmentation}
\label{sub:sec:learnable:feature:augmentation}
This technique allows us to learn node feature augmentations.
Given a graph ${G = (\mathbf{A}, \mathbf{X})}$, we apply a learnable feature augmentation 
$A_{f_\Theta}$ on $G$ as:
\begin{equation*}
    A_{f_\Theta}(G) = (\mathbf{A}, f_{\Theta}(\mathbf{X}))
\end{equation*}
and we model $f$ as a feed-forward neural network (FFN). 
Two separate learnable functions $f_{\Theta_1}$ and  $f_{\Theta_2}$, parameterized by 
$\Theta_1$ and $\Theta_2$ compute two augmented views of $\mathbf{X}$ as:
\begin{align*}
    \mathbf{X_1} &= f_{\Theta_1}(\mathbf{X}) \in \mathbb{R}^{B \times D_1}\\
    \mathbf{X_2} &= f_{\Theta_2}(\mathbf{X}) \in \mathbb{R}^{B \times D_1}
\end{align*}
where
\[\Theta_1 = \lbrace \mathbf{W}_1^{(l)} : l = 1, \ldots, L \rbrace,  
\Theta_2 = \lbrace \mathbf{W}_2^{(l)} : l = 1, \ldots, L \rbrace\]
are set of weights, and $\mathbf{W}_1^{(l)}$ or $\mathbf{W}_2^{(l)}$ are the weight matrices of the $l$--th layer 
of the FFNs, $B$ is batch-size, $D_1$ is the augmentation dimension, and $L$ is the number of layers of the FFNs.
Figure~\ref{fig:architecture} (a) shows~\model's architecture based on learnable feature augmentation.

\subsubsection{Learnable Topology Augmentation}
\label{sub:sec:learnable:topology:augmentation}
Studies have shown that using diffusion-based high-order networks improves the performance of GNNs~\cite{klicpera2019diffusion}.
Consequently, high-order networks have been used for augmentation in SSGRL.
This study proposes a complementary approach that learns the $K$--order relation between nodes.
That is, we apply $A_{t_\Phi'}$ on $G$ as:
\[{A_{t_\Phi'}(G) = (t_\Phi'(\mathbf{A}), \mathbf{X})}\]
to obtain a high-order network
\[\mathbf{A}' = t_\Phi'(\mathbf{A})\]

First, we learn a latent representation, ${\mathbf{H} \in \mathbb{R}^{B \times D_1}}$, which encodes high-order ($K$--hop) signal.
To this end, we employ a GNN, as GNNs enable us to receive a signal from 
$K$--hop neighbors similar to static diffusion algorithms, such as personalized PageRank and heat kernel.
Hence, for each node $i$, a GNN $h_\Phi$, parameterized by $\Phi$ is used to learn high-order feature vector $\mathbf{H}[i] = \mathbf{h}_i$ and $\mathbf{H}$ is computed as:
\begin{equation*}
    \mathbf{H} = h_\Phi(\mathbf{A}, \mathbf{X})
\end{equation*}
where ${\Phi = \lbrace \mathbf{W}^{(l)}: l = 1, \ldots, K \rbrace}$,
$\mathbf{W}^{(l)}$ is the weight matrix of the $l$-th layer of the GNN, and $K$ is the number of layers.
 
We obtain two topological views, which are $\mathbf{A}' = t_{\Phi'}(\mathbf{A})$ and $\mathbf{A} = t_{\Phi}(\mathbf{A})$,
where $t_{\Phi}$ is simply an identity augmentor.
The high-order network $\mathbf{A}'$ is constructed based on the high-order features as
\begin{equation*}
    \mathbf{A}' = t_{\Phi}'(\mathbf{A}) = g(\mathbf{H}, \mathbf{H})
\end{equation*}
The entry $\mathbf{a}'_{ij}$ is computed as
\[
\mathbf{a}'_{ij} = 
\begin{cases}
    g(\mathbf{h}_i, \mathbf{h}_j),& \text{if }  g(\mathbf{h}_i, \mathbf{h}_j) > \mathbb E_{k\in V}[g(\mathbf{h}_i, \mathbf{h}_k)]\\
    0,              & \text{otherwise}
\end{cases}
\]
and $g$ is defined as
\begin{equation*}
    g(\mathbf{h}_i, \mathbf{h}_j) = \mathbf{h}^T_i \cdot \mathbf{h}_j
\end{equation*}
\model's architecture based on this augmentation technique is depicted in Fig.~\ref{fig:architecture}(b).


\subsection{Encoding}
\label{sub:sec:encoding}
After generating two views of the graph, either using the feature or topology augmentor, we feed each view
independently to a shared GNN encoder, $h_\Theta$, to learn a latent graph representation.
For brevity, regardless of the augmentor, we refer to the views in the augmentation space as 
$G_1 = (\mathbf{A}_1, \mathbf{X}_1)$ and $G_2 = (\mathbf{A}', \mathbf{X}_2)$.
The next task is to learn two latent representations ${\mathbf{Z_1} \in \mathbb{R}^{B \times D}}$ and 
${\mathbf{Z_2}  \in \mathbb{R}^{B \times D}}$ that encode the two views, where $D$ is the number of latent dimensions.
We achieve this by using a shared GNN, $h_\Theta$, as:
\[
\mathbf{Z_1} = h_\Theta(\mathbf{A}, \mathbf{X}_1)
\]
\[
\mathbf{Z_2} = h_\Theta(\mathbf{A}', \mathbf{X}_2)
\]
For full-batch training, the batch axis becomes $N$ instead of $B$.
Henceforth though, we assume a mini-batch training.

\subsection{Training}
\label{sub:sec:training} 
Generally, in SSGRL, we want the latent representations $\mathbf{Z}_1$ and $\mathbf{Z}_2$ of nodes to be 
invariant to the perturbations.
For this reason, we want to maximize the agreement (similarity) between $\mathbf{Z}_1$ and $\mathbf{Z}_2$.
Minimizing the L-2 distance is commonly used for this purpose; thus, we use the same strategy.
We closely follow a similar formulation as~\cite{bardes2021vicreg} and define a term called invariance based on the L-2 distance as:
\begin{equation}\label{eq:objective:base}
    \texttt{inv} = \vert \vert \mathbf{Z}_1 - \mathbf{Z}_2 \vert \vert_F
\end{equation}
Nonetheless, this has a trivial solution that collapses the representations.
Several strategies, mostly engineering tricks, have been used to prevent this collapse~\cite{DBLP:journals/corr/abs-2006-07733,thakoor2021bootstrapped,DBLP:journals/corr/abs-2103-14958}.
Instead, we use a principled approach inspired by a recent method~\cite{bardes2021vicreg} proposed for visual representation.
That is, we add two regularization terms called variance and covariance regularizations.

The variance regularization is defined as
\begin{equation}\label{eq:variance:regularization}
    v(\mathbf{Z}) = \frac{1}{D} \sum_{j=1}^D \max(0, 1 - \sqrt{Var(\mathbf{z}_{:j}) + \epsilon})
\end{equation}
and it constrains each dimension of the latent representation to have a variance of 1; as a result 
prevents data points from collapsing into a subspace.

The covariance term is defined as 
\begin{equation}\label{eq:covariance:regularization}
    c(\mathbf{Z}) = \frac{1}{D} \sum_{i\neq j} \Big[\frac{ \mathbf{\bar{Z}}^T \mathbf{\bar{Z}}}{B-1}\Big]^2_{i, j}
\end{equation}
and it is the sum of the squared-off diagonal elements of the covariance matrix $\mathbf{\bar{Z}}^T \mathbf{\bar{Z}}$,
where $\mathbf{\bar{Z}}$ is the mean centered representation.
The covariance term is normalized and scaled with respect to $B$ and $D$, respectively.
As a result of constraining the off-diagonal elements of the covariance matrix to be zero, this regularization ensures 
that each dimension is independent of each other, consequently preventing the dimensions from collapsing.
Finally, we define the regularization on the latent space as:
\begin{equation}\label{eq:representation:regularization}
    R_{\mathbf{Z}_1,\mathbf{Z}_2} = \beta * (v(\mathbf{Z}_1) + v(\mathbf{Z}_2)) +
                                    \gamma * (c(\mathbf{Z}_1) + c(\mathbf{Z}_2))
\end{equation}


Furthermore, the augmentation models can collapse, 
that is, ${f_{\Theta_1} = f_{\Theta_2}}$; albeit, empirically, this has not been observed. 
Since the regularizations mentioned above on the latent space only ensure that neither $\mathbf{Z}_1$ nor 
$\mathbf{Z}_2$ collapse along the batch or the dimension axes.
Thus, to prevent model collapse, we define a model regularization term as:

\begin{equation}\label{eq:model:regularization}
R_{\Theta_1, \Theta_2} = \sum_{\mathbf{W}_l} \vert \vert \mathbf{W}_l \mathbf{W}_l^T - \mathbf{I} \vert \vert_F
\end{equation}
where $\mathbf{W}_l = \big[\begin{smallmatrix}
  \mathbf{W}_1^{(l)} \\
  \mathbf{W}_2^{(l)}
\end{smallmatrix}\big]$ is a vertical stacking of  $\mathbf{W}_1^{(l)} \in \Theta_1$ and $\mathbf{W}_2^{(l)} \in \Theta_2$;
recall that $\mathbf{W}^{(l)}_1 $ and $\mathbf{W}^{(l)}_2$ are weights of the $l^{th}$ layer of a FFN.
For any two row indices $i$ and $j$  of $\mathbf{W}_l$, where ${i \neq j}$, the model regularization encourages each row 
vector $\mathbf{W}_l[i]$ to be orthogonal to any other vector $\mathbf{W}_l[j]$.
Consequently, ${\nexists i: \mathbf{W}^{(l)}_{1}[i] =  \mathbf{W}^{(l)}_{2}[i]} \Rightarrow {\mathbf{W}^{(l)}_1 \neq \mathbf{W}^{(l)}_2}$.

The overall training cost function is then defined as:
\begin{equation}\label{eq:training:cost:function}
\mathcal{L}_{\Psi} =  \alpha * \texttt{inv} + R_{\mathbf{Z}_1,\mathbf{Z}_2}  + \lambda * R_{\Theta_1, \Theta_2}
\end{equation}

where $\Psi$ is the set of all model parameters, that is, ${\Psi = \lbrace \Theta, \Theta_1, \Theta_2 \rbrace}$
for the model based on feature augmentation and ${\Psi = \lbrace \Theta, \Phi \rbrace}$ for the model based on
topology augmentation.
The coefficients $\alpha, \beta, \gamma$, and $\lambda$ control the contribution of the different cost function terms.
In most cases, we have observed that setting these values to just one works well.
\begin{table*}
\centering
\caption{Summary of the datasets used for node classification experiment. BCC, MCC, and MLC refer to the  
classification task, which is binary, multi-class, and multilabel classification, respectively}
\label{tbl:node:classification:datasets}
\begin{tabular}{|l|l|l|l|l|l|l||l|l|}
\hline
\textbf{Dataset} & MAG-CS & AmazonPhoto & PubMed & GitHub & WikiCS & Dezer & Yelp & Reddit \\ \hline
N & 18,333 & 7,650 & 19,717 & 37,700 & 11,701 & 28,281 & 716,847 & 232,965 \\ \hline
M & 163,788 & 238,162 & 88,648 & 578,006 & 297,110 & 185,504 & 13,954,819 & 114,615,892 \\ \hline
F & 6,805 & 745 & 500 & 128 & 300 & 128 & 300 & 602 \\ \hline
\#classes & 15 (MCC) & 8 (MCC) & 3 (MCC) & 2 (BCC) & 15 (MCC) & 2 (BCC) & 100 (MLC) & 41 (MCC) \\ \hline
\end{tabular}
\end{table*}

\begin{table}
\centering
\caption{Summary of the dataset used for graph property prediction task,  \#G is the number of graphs and 
$\tilde{N}$ and $\tilde{M}$ are the average number of nodes and edges in each graph, respectively.}
\label{tbl:graph:property:prediction:datasets}
\begin{tabular}{|l|l|l|l|l|l|}
\hline
Datasets & \#G &  $\tilde{N}$ & $\tilde{M}$ & F & \#classes \\ \hline
DD & 1178 & 284.32 & 715.66 & 89 & 2 \\ \hline
ENZYMES & 600 & 32.63 & 62.14 & 21 & 6 \\ \hline
PROTEINS & 1113 & 39.06 & 72.82 & 4 & 2 \\ \hline
NCI1 & 4110 & 29.87 & 32.3 & 1 & 2 \\ \hline
IMDB-B & 1000 & 19.77 & 96.53 & 5 & 2 \\ \hline
IMDB-M & 1500 & 13 & 65.94 & 5 & 3 \\ \hline
REDDIT & 2000 & 429.63 & 497.75 & 5 & 2 \\ \hline
COLLAB & 5000 & 74.49 & 2457.78 & 5  & 3 \\ \hline
\end{tabular}
\end{table}

If one desires to reduce the number of hyper-parameters, we provide an alternative formulation for 
Eq.~\ref{eq:representation:regularization}. 
By following the same formulation as Eq.~\ref{eq:model:regularization}, an alternative formulation 
that is inspired by  Laplacian Eigenmaps~\cite{6789755} is defined:
\begin{equation}\label{eq:orthonormality:regularization}
    R_{\mathbf{Z}_1,\mathbf{Z}_2} =  \gamma * (\vert \vert \mathbf{\tilde{Z}_1}\mathbf{\tilde{Z}_1}^T - \mathbf{I} \vert \vert_F +
                                              \vert \vert \mathbf{\tilde{Z}_2}\mathbf{\tilde{Z}_2}^T - \mathbf{I} \vert \vert_F)
\end{equation}  
where $\mathbf{I} \in \mathbf{R}^{B \times B}$ is an identity matrix and each row vector $\mathbf{\tilde{z}_i}$ of $\mathbf{\tilde{Z}}$ 
is a unit vector.
We refer to Eq.~\ref{eq:orthonormality:regularization} as an orthogonality regularization, and empirically it performs similarly to Eq.~\ref{eq:representation:regularization}.
The main disadvantage of Eq.~\ref{eq:orthonormality:regularization} is that it could be expensive for full-batch GNNs, where $B=N$.


Note that~\model~is jointly optimized on both the augmenter and encoder parameters.
As a result, the learned augmentations are governed by the inherent signal in the data.

\section{Empirical Evaluation}
\label{sec:experiments}

We validate the proposed method on node classification (NC) and graph property prediction (GPP) tasks.
In the former case, the prediction is at a node level, and for the latter, it is at a graph level. Additional thorough analysis of the experiments and the running times of the methods and  are included in the appendix. \footnote{\url{https://github.com/AhmedESamy/dsgrl/blob/main/Appendix.pdf}}

\subsection{Datasets}
The datasets are 8 for NC and 8 for GPP, and a summary is provided in Tables~\ref{tbl:node:classification:datasets} and
~\ref{tbl:graph:property:prediction:datasets}.

\subsubsection{NC Datasets}
\begin{itemize}
    \item Citation Networks (PubMed): Paper-to-paper citation networks, and we classify papers into different subjects~\cite{hamilton2018inductive}.
    \item Co-Author Networks (MAG-CS): Author collaboration network from Microsoft Academic Graph, and the task is to predict the active field of authors~\cite{shchur2019pitfalls}.
    \item Co-Purchased Products Network (AmazonPhoto): Co-purchased products from Amazon Photo Category, and the task is to predict the refined categories~\cite{shchur2019pitfalls}.
    \item Wikipedia (WikiCS): Wikipedia hyperlinks between Computer Science articles, and we classify articles into branches of CS~\cite{shchur2019pitfalls}.
    \item Social (Facebook, GitHub, Reddit, and Yelp):
    Facebook contains a page-to-page graph of verified Facebook sites, and we want to classify pages into their categories~\cite{rozemberczki2021multiscale}. 
    GitHub contains the social network of developers, and we want to classify developers as web or machine learning developers~\cite{rozemberczki2021multiscale}. Yelp is also the social network of Yelp users, and we predict the business categories each user has reviewed. For Reddit, we predict the subreddits (communities) of user posts~\cite{hamilton2018inductive,zeng2020graphsaint}. 
\end{itemize}
\subsubsection{GPP Datasets}
\begin{itemize}
    \item Chemical Datasets (DD, NCI1, PROTEINS, ENZYMES)~\cite{Morris+2020} that represent protein interaction or molecular graphs.
    The task is to predict different properties of molecules or macromolecules.
    \item Social Datasets (IMDB-BINARY, IMDB-MULTI, REDDIT-BINARY, COLLAB)~\cite{Morris+2020} that represent the collaboration between users in different ego-networks. The task is to predict the class of the ego-networks.
\end{itemize}

\subsection{Baselines}
\begin{table*}[ht!]
\centering
\caption{Results of the NC experiment for the small datasets. \oor~corresponds to out-of-resource (GPU Memory) Random-F and Random-T are untrained variants of DSGRL based on feature and toplogy augmentions, respectively.}
\label{tbl:node:classification:results:on:small:datasets}
\begin{tabular}{|ll|llllll|}
\hline
\multicolumn{2}{|l|}{\multirow{2}{*}{Methods}} & \multicolumn{6}{c|}{Datasets} \\ \cline{3-8} 
\multicolumn{2}{|l|}{} & \multicolumn{1}{l|}{MAG-CS} & \multicolumn{1}{l|}{AmazonPhoto} & \multicolumn{1}{l|}{PubMed} & \multicolumn{1}{l|}{GitHub} & \multicolumn{1}{l|}{WikiCS} & Deezer \\ \hline

\multicolumn{1}{|c|}{\multirow{4}{*}{Ours}} & Random-F & \multicolumn{1}{l|}{80.2$\pm$0.1} & \multicolumn{1}{l|}{82.33$\pm$0.1} & \multicolumn{1}{l|}{75.5$\pm$0.1} & \multicolumn{1}{l|}{82.5$\pm$0.1} & \multicolumn{1}{l|}{63.4$\pm$0.3} & 55.4$\pm$0.2 \\ \cline{2-8} 
\multicolumn{1}{|l|}{} & Random-T & \multicolumn{1}{l|}{69.8$\pm$0.1} & \multicolumn{1}{l|}{80.6$\pm$0.3} & \multicolumn{1}{l|}{71.4$\pm$0.1} & \multicolumn{1}{l|}{78.8$\pm$0.1} & \multicolumn{1}{l|}{63.6$\pm$0.5} & 56.0$\pm$0.2 \\ \cline{2-8} 
\multicolumn{1}{|l|}{} & Feature & \multicolumn{1}{l|}{91.4$\pm$0.1} & \multicolumn{1}{l|}{\textbf{90.6$\pm$0.1}} & \multicolumn{1}{l|}{82.4$\pm$0.2} & \multicolumn{1}{l|}{\textbf{84.9$\pm$0.5}} & \multicolumn{1}{l|}{\textbf{74.2$\pm$0.1}} & 58.2$\pm$0.1 \\ \cline{2-8}
\multicolumn{1}{|l|}{} & Topology & \multicolumn{1}{l|}{\textbf{92.9$\pm$0.1}} & \multicolumn{1}{l|}{89.7$\pm$0.1} & \multicolumn{1}{l|}{\textbf{83.8$\pm$0.1}} & \multicolumn{1}{l|}{83.0$\pm$0.3} & \multicolumn{1}{l|}{73.6$\pm$0.1} & \textbf{59.3$\pm$0.1} \\ \hline

\multicolumn{1}{|l|}{\multirow{4}{*}{\begin{tabular}[c]{@{}c@{}}Self-Supervised \\ (Baselines)\end{tabular}}} & \dgi & \multicolumn{1}{l|}{91.1$\pm$0.2} & \multicolumn{1}{l|}{89.0$\pm$0.6} & \multicolumn{1}{l|}{78.6$\pm$0.5} & \multicolumn{1}{l|}{79.1$\pm$0.7} & \multicolumn{1}{l|}{73.6$\pm$0.4} & 55.2$\pm$0.6 \\ \cline{2-8} 
\multicolumn{1}{|l|}{} & \mvgrl & \multicolumn{1}{l|}{88.2$\pm$0.1} & \multicolumn{1}{l|}{87.2$\pm$0.1} & \multicolumn{1}{l|}{77.0$\pm$0.3} & \multicolumn{1}{l|}{79.8$\pm$0.1} & \multicolumn{1}{l|}{61.7$\pm$0.1} & \oor \\  \cline{2-8} 
\multicolumn{1}{|l|}{} & \gca & \multicolumn{1}{l|}{91.0$\pm$0.4} & \multicolumn{1}{l|}{86.0$\pm$1.1} & \multicolumn{1}{l|}{\textbf{83.8$\pm$0.2}} & \multicolumn{1}{l|}{\oor} & \multicolumn{1}{l|}{72.9$\pm$0.6} & \oor \\ \cline{2-8} 
\multicolumn{1}{|l|}{} & \byol & \multicolumn{1}{l|}{90.7$\pm$0.3} & \multicolumn{1}{l|}{90.3$\pm$0.5} & \multicolumn{1}{l|}{82.4$\pm$0.4} & \multicolumn{1}{l|}{81.3$\pm$0.4} & \multicolumn{1}{l|}{73.8$\pm$0.7} & 58.2$\pm$0.7 \\ \hline \hline

\multicolumn{1}{|c|}{\multirow{3}{*}{\begin{tabular}[c]{@{}c@{}}Semi-Supervised\\ (References)\end{tabular}}} & \gcn & \multicolumn{1}{l|}{91.7$\pm$0.3} & \multicolumn{1}{l|}{92.0$\pm$0.4} & \multicolumn{1}{l|}{85.4$\pm$0.4} & \multicolumn{1}{l|}{84.1$\pm$0.3} & \multicolumn{1}{l|}{76.7$\pm$0.6} & 59.7$\pm$0.6 \\ \cline{2-8} 
\multicolumn{1}{|l|}{} & \gat & \multicolumn{1}{l|}{91.3$\pm$0.1} & \multicolumn{1}{l|}{92.3$\pm$0.5} & \multicolumn{1}{l|}{84.7$\pm$0.1} & \multicolumn{1}{l|}{85.5$\pm$0.3} & \multicolumn{1}{l|}{77.3$\pm$0.5} & 59.5$\pm$0.6 \\ \cline{2-8} 
\multicolumn{1}{|l|}{} & \sage & \multicolumn{1}{l|}{91.6$\pm$0.3} & \multicolumn{1}{l|}{92.4$\pm$0.4} & \multicolumn{1}{l|}{84.5$\pm$0.4} & \multicolumn{1}{l|}{84.6$\pm$0.4} & \multicolumn{1}{l|}{77.4$\pm$0.6} & 61.9$\pm$0.6 \\ \hline
\end{tabular}
\end{table*}

\begin{table*}[ht!]
\centering
\caption{Results of the NC experiment for two of the large-scale datasets. 
We only include semi-supervised and scalable GNN architectures for this experiment as the full-batch ones do not fit in GPU memory. 
In addition, all the SSGRL baselines throw an out-of-memory error.}
\label{tbl:node:classification:results:on:large:datasets}
\resizebox{\columnwidth}{!}{
\begin{tabular}{|l|c|c|}
\hline
\multicolumn{1}{|c|}{\multirow{2}{*}{Methods}} & \multicolumn{2}{c|}{Datasets} \\ \cline{2-3} 
\multicolumn{1}{|c|}{}      & Yelp (ROC-AUC)  & Reddit (Accuracy) \\ \hline
        \cluster~(semi)           & 78.2 & 95.3  \\ \hline
        \saint~(semi)           & 75.6 & 95.7  \\ \hline
        \pprgo~(semi)           & 77.7  & 91.8   \\ \hline
        \model~(Random-F)             & 72.6 & 82.3  \\ \hline
        \model~(Feature)             & 75.2 & 89.3  \\ \hline 
\end{tabular}
}
\end{table*}

\begin{table*}[t!]
\centering
\caption{Results of the NC experiment for heterogeneous graph.}
\label{tbl:node:classification:heterogeneous}
\resizebox{\textwidth}{!}{
\begin{tabular}{|c|c|cccccc||ccc|}
\hline
\multirow{2}{*}{Metrics} & \multirow{2}{*}{Train \%} & \multicolumn{4}{c|}{Unsupervised/Self-supervised (Baselines)} &
\multicolumn{2}{c||}{Ours} & \multicolumn{3}{c|}{Semi-Supervised (References)} \\ \cline{3-11} 
 &  & \multicolumn{1}{c|}{\nodetovec~\cite{grover2016node2vec}} & \multicolumn{1}{c|}{\esim~\cite{https://doi.org/10.48550/arxiv.1610.09769}} & \multicolumn{1}{c|}{\metapathtovec~\cite{10.1145/3097983.3098036}} & \multicolumn{1}{c|}{\herec~\cite{https://doi.org/10.48550/arxiv.1711.10730}} &
 \multicolumn{1}{l|}{Random-F} & \model & \multicolumn{1}{c|}{\gat} & \multicolumn{1}{c|}{\han~\cite{10.1145/3308558.3313562}} & \magnn~\cite{Fu_2020} \\ \hline
\multirow{4}{*}{Macro-F1} & 20 & \multicolumn{1}{c|}{49.00} & \multicolumn{1}{c|}{48.37} & \multicolumn{1}{c|}{46.05} & \multicolumn{1}{c|}{45.61} & \multicolumn{1}{l|}{41.58} & \textbf{53.14} & \multicolumn{1}{c|}{53.64} & \multicolumn{1}{c|}{56.19} & 59.35 \\ \cline{2-11} 
 & 40 & \multicolumn{1}{c|}{50.63} & \multicolumn{1}{c|}{50.09} & \multicolumn{1}{c|}{47.57} & \multicolumn{1}{c|}{46.80} & \multicolumn{1}{l|}{44.84} & \textbf{54.90} & \multicolumn{1}{c|}{55.50} & \multicolumn{1}{c|}{56.15} & 60.27 \\ \cline{2-11} 
 & 60 & \multicolumn{1}{c|}{51.65} & \multicolumn{1}{c|}{51.45} & \multicolumn{1}{c|}{48.17} & \multicolumn{1}{c|}{46.84} & \multicolumn{1}{l|}{44.12} & \textbf{56.25} & \multicolumn{1}{c|}{56.46} & \multicolumn{1}{c|}{57.29} & 60.66 \\ \cline{2-11} 
 & 80 & \multicolumn{1}{c|}{51.49} & \multicolumn{1}{c|}{51.37} & \multicolumn{1}{c|}{49.99} & \multicolumn{1}{c|}{47.73} & \multicolumn{1}{l|}{45.20} & \textbf{60.28} & \multicolumn{1}{c|}{57.43} & \multicolumn{1}{c|}{58.51} & 61.44 \\ \hline
\multirow{4}{*}{Micro-F1} & 20 & \multicolumn{1}{c|}{49.94} & \multicolumn{1}{c|}{49.32} & \multicolumn{1}{c|}{47.22} & \multicolumn{1}{c|}{46.23} & \multicolumn{1}{l|}{42.44} & \textbf{53.35} & \multicolumn{1}{c|}{53.64} & \multicolumn{1}{c|}{56.32} & 59.60 \\ \cline{2-11} 
 & 40 & \multicolumn{1}{c|}{51.77} & \multicolumn{1}{c|}{51.21} & \multicolumn{1}{c|}{48.17} & \multicolumn{1}{c|}{47.89} & \multicolumn{1}{l|}{45.57} & \textbf{54.89} & \multicolumn{1}{c|}{55.56} & \multicolumn{1}{c|}{57.32} & 60.50 \\ \cline{2-11} 
 & 60 & \multicolumn{1}{c|}{52.79} & \multicolumn{1}{c|}{52.53} & \multicolumn{1}{c|}{49.17} & \multicolumn{1}{c|}{48.19} & \multicolumn{1}{l|}{44.61} & \textbf{56.32} & \multicolumn{1}{c|}{56.47} & \multicolumn{1}{c|}{58.42} & 60.88 \\ \cline{2-11} 
 & 80 & \multicolumn{1}{c|}{52.72} & \multicolumn{1}{c|}{52.54} & \multicolumn{1}{c|}{50.50} & \multicolumn{1}{c|}{49.11} & \multicolumn{1}{l|}{45.55} & \textbf{60.05} & \multicolumn{1}{c|}{57.40} & \multicolumn{1}{c|}{59.24} & 61.53 \\ \hline
\end{tabular}
}
\end{table*}
\begin{table*}
\caption{Results for graph property prediction experiment. We report the classification accuracy of three groups of methods: Semi-Supervised (References), Self-Supervised (Baselines), and the variants of our method. Bold indicates the best-performing method.}
\label{tbl:graph:property:prediction:results}
\resizebox{\textwidth}{!}{
\begin{tabular}{|ll|llllllll|}
\hline
\multicolumn{2}{|c|}{\multirow{3}{*}{Methods}} & \multicolumn{8}{c|}{Datasets} \\ \cline{3-10} 
\multicolumn{2}{|l|}{} & \multicolumn{4}{c|}{Chemical} & \multicolumn{4}{c|}{Social} \\ \cline{3-10} 
\multicolumn{2}{|l|}{} & \multicolumn{1}{l|}{DD} & \multicolumn{1}{l|}{NCI1} & \multicolumn{1}{l|}{PROTEINS} & \multicolumn{1}{l|}{ENZYMES} & \multicolumn{1}{l|}{IMDB-B} & \multicolumn{1}{l|}{IMDB-M} & \multicolumn{1}{l|}{REDDIT} & COLLAB \\ \hline

\multicolumn{1}{|c|}{\multirow{4}{*}{Ours}} & Random-F & \multicolumn{1}{l|}{77.0$\pm$2.9} & \multicolumn{1}{l|}{67.2$\pm$1.9} & \multicolumn{1}{l|}{74.4$\pm$3.7} & \multicolumn{1}{l|}{36.0$\pm$5.7} & \multicolumn{1}{l|}{70.7$\pm$4.0} & \multicolumn{1}{l|}{46.5$\pm$3.6} & \multicolumn{1}{l|}{76.7$\pm$2.1} & 66.3$\pm$2.7 \\ \cline{2-10} 
\multicolumn{1}{|l|}{} & Ramdom-T & \multicolumn{1}{l|}{75.6$\pm$2.3} & \multicolumn{1}{l|}{71.0$\pm$1.9} & \multicolumn{1}{l|}{73.0$\pm$1.8} & \multicolumn{1}{l|}{36.0$\pm$5.6} & \multicolumn{1}{l|}{67.0$\pm$4.7} & \multicolumn{1}{l|}{43.3$\pm$3.9} & \multicolumn{1}{l|}{69.4$\pm$1.1} & 66.5$\pm$1.5 \\ \cline{2-10} 
\multicolumn{1}{|l|}{} & Feature & \multicolumn{1}{l|}{\textbf{78.0$\pm$2.9}} & \multicolumn{1}{l|}{75.0$\pm$2.2} & \multicolumn{1}{l|}{\textbf{75.6$\pm$3.4}} & \multicolumn{1}{l|}{\textbf{54.0$\pm$3.7}} & \multicolumn{1}{l|}{\textbf{71.9$\pm$3.4}} & \multicolumn{1}{l|}{\textbf{50.3$\pm$3.4}} & \multicolumn{1}{l|}{\textbf{78.3$\pm$2.1}} & 69.6$\pm$1.4 \\ \cline{2-10} 
\multicolumn{1}{|l|}{} & Topology & \multicolumn{1}{l|}{75.8$\pm$3.0} & \multicolumn{1}{l|}{72.8$\pm$1.9} & \multicolumn{1}{l|}{74.4$\pm$4.5} & \multicolumn{1}{l|}{37.8$\pm$5.7} & \multicolumn{1}{l|}{71.7$\pm$3.6} & \multicolumn{1}{l|}{\textbf{50.3$\pm$3.3}} & \multicolumn{1}{l|}{\textbf{78.3$\pm$2.3}} & \textbf{70.0$\pm$2.1} \\ \hline 
\multicolumn{1}{|l|}{\multirow{3}{*}{\begin{tabular}[c]{@{}c@{}}Self-Supervised\\ (Baselines)\end{tabular}}} & \adgcl & \multicolumn{1}{l|}{69.9$\pm$3.8} & \multicolumn{1}{l|}{67.7$\pm$1.2} & \multicolumn{1}{l|}{71.2$\pm$2.28} & \multicolumn{1}{l|}{20.5$\pm$3.6} & \multicolumn{1}{l|}{69.1$\pm$3.4} & \multicolumn{1}{l|}{41.7$\pm$2.2} & \multicolumn{1}{l|}{70.8$\pm$2.6} & 67.7$\pm$2.5 \\ \cline{2-10} 
\multicolumn{1}{|l|}{} & \graphcl & \multicolumn{1}{l|}{76.4$\pm$2.7} & \multicolumn{1}{l|}{\textbf{75.2$\pm$1.3}} & \multicolumn{1}{l|}{73.7$\pm$5.0} & \multicolumn{1}{l|}{25.3$\pm$6.0} & \multicolumn{1}{l|}{\textbf{71.9$\pm$4.8}} & \multicolumn{1}{l|}{47.2$\pm$4.1} & \multicolumn{1}{l|}{78.1$\pm$1.9} & 69.24$\pm$1.1 \\ \cline{2-10} 
\multicolumn{1}{|l|}{} & \infograph & \multicolumn{1}{l|}{74.8$\pm$4.0} & \multicolumn{1}{l|}{73.4$\pm$2.1} & \multicolumn{1}{l|}{73.8$\pm$3.6} & \multicolumn{1}{l|}{30.1$\pm$5.1} & \multicolumn{1}{l|}{71.5$\pm$2.5} & \multicolumn{1}{l|}{47.8$\pm$4.0} & \multicolumn{1}{l|}{73.5$\pm$2.9} & 64.1$\pm$1.2 \\ \hline \hline

\multicolumn{1}{|l|}{\multirow{5}{*}{\begin{tabular}[c]{@{}c@{}} Semi-Supervised\\ (References)\end{tabular}}} & \dgcnn & \multicolumn{1}{l|}{76.6 -/+ 4.3} & \multicolumn{1}{l|}{76.4$\pm$1.7} & \multicolumn{1}{l|}{72.9$\pm$3.5} & \multicolumn{1}{l|}{38.9$\pm$5.7} & \multicolumn{1}{l|}{69.2$\pm$3.0} & \multicolumn{1}{l|}{45.6$\pm$3.4} & \multicolumn{1}{l|}{87.8$\pm$2.5} & 71.2$\pm$1.9 \\ \cline{2-10} 
\multicolumn{1}{|l|}{} & \diffpool & \multicolumn{1}{l|}{75.0$\pm$3.5} & \multicolumn{1}{l|}{76.9$\pm$1.9} & \multicolumn{1}{l|}{73.7$\pm$3.5} & \multicolumn{1}{l|}{59.5$\pm$5.6} & \multicolumn{1}{l|}{68.4$\pm$3.3} & \multicolumn{1}{l|}{45.6$\pm$3.4} & \multicolumn{1}{l|}{89.1$\pm$1.6} & 68.9$\pm$2.0 \\ \cline{2-10} 
\multicolumn{1}{|l|}{} & \ecc & \multicolumn{1}{l|}{72.6$\pm$4.1} & \multicolumn{1}{l|}{76.2$\pm$1.4} & \multicolumn{1}{l|}{72.3$\pm$3.4} & \multicolumn{1}{l|}{29.5$\pm$8.2} & \multicolumn{1}{l|}{67.7$\pm$2.8} & \multicolumn{1}{l|}{43.5$\pm$3.1} & \multicolumn{1}{l|}{OOR} & OOR \\ \cline{2-10} 
\multicolumn{1}{|l|}{} & \gin & \multicolumn{1}{l|}{75.3$\pm$2.9} & \multicolumn{1}{l|}{80.0$\pm$1.4} & \multicolumn{1}{l|}{73.3$\pm$4.0} & \multicolumn{1}{l|}{59.6$\pm$4.5} & \multicolumn{1}{l|}{71.2$\pm$3.9} & \multicolumn{1}{l|}{48.5$\pm$3.3} & \multicolumn{1}{l|}{89.9$\pm$1.9} & 75.6$\pm$2.3 \\ \cline{2-10} 
\multicolumn{1}{|l|}{} & \sage & \multicolumn{1}{l|}{72.9$\pm$2.0} & \multicolumn{1}{l|}{76.0$\pm$1.8} & \multicolumn{1}{l|}{73.0$\pm$4.5} & \multicolumn{1}{l|}{58.2$\pm$6.0} & \multicolumn{1}{l|}{68.8$\pm$4.5} & \multicolumn{1}{l|}{47.6$\pm$3.5} & \multicolumn{1}{l|}{84.3$\pm$1.9} & 73.9$\pm$1.7 \\ \hline
\end{tabular}
}
\end{table*}

We compare our method against strong self-supervised baselines.
As a result of a plethora of related methods, baselines are selected either based on their popularity or if they are current SOTA methods that outperform existing methods.
Hence, for the NC task, we select
\dgi~\cite{velickovic2018deep}, a method that uses corruption based on permutation of node features and topology, 
\mvgrl~\cite{DBLP:journals/corr/abs-2006-05582}, which augments the topology using high-order networks
obtained through a diffusion process, and
\gca~\cite{10.1145/3442381.3449802}~ is a method based on adaptive edge removal and feature masking augmentations. 
All of these use a contrastive architecture using mutual information maximization with negative sampling. Furthermore, we include a contrastive architecture based on asymmetry called~\byol~\cite{DBLP:journals/corr/abs-2006-07733} for completeness. 
Although it was initially proposed for visual representations, recent studies~\cite{thakoor2021bootstrapped,DBLP:journals/corr/abs-2103-14958} have extended it for GRL.
We use the best augmentations reported in these papers.

As there are several studies for the GPP task, we select strong representative baselines, which are
\graphcl~\cite{DBLP:journals/corr/abs-2106-07594,DBLP:journals/corr/abs-2010-13902}, \adgcl~\cite{suresh2021adversarial}, and ~\infograph~\cite{sun2020infograph}. For example, SimGrace \cite{xia2022simgrace} is shown to give sub-optimal performance compared to GraphCL \cite{DBLP:journals/corr/abs-2106-07594} as reported in \cite{li2023multi, tangraph}, therefore, we only include the results for GraphCL in our experiments.~\graphcl~learns augmentations from a set of augmentations whereas~\adgcl~learns to drop edges using adversarial training.
~\infograph~learns by maximizing the mutual information between graph and patch (subgraphs, node, edges) level representations. 

Furthermore, we include untrained variants of~\model~as suggested in~\cite{trivedi2021augmentations}.
We refer to them as Random-F and Random-T to denote the feature and topology augmentation-based architectures, respectively.

All the above baselines are self-supervised and unsupervised methods.
We also include semi-supervised methods; however, they are included just for reference and 
not comparison.


\subsection{Node Classification on Homogeneous Graphs}
\label{sub:sec:node:classification:experiment}
Following the recommended evaluation protocol for node classification~\cite{shchur2019pitfalls}, we create ten splits for each dataset where there is no public split. 
We use the linear evaluation protocol to quantify the quality of the representations obtained from the SSGRL methods, where we first train each SSGRL method on each split with no labels.
Then, for each split, we set 5\% and 15\% as training and validation splits used for model selection and 80\% for testing using a Linear (Logistic) classifier.
Model selection is carried out using only 1 of the ten splits, and for a fair comparison, it is done for all the baselines.
We tune all the hyper-parameters using Bayesian optimization~\footnote{Bayesian optimization using OPTUNA: \url{https://optuna.org/}}.
In addition, for the baselines, the size of the representation dimension, $D$, is 128; for our model, it is 64, since we concatenate $\mathbf{Z}_1$ and $\mathbf{Z}_2$.
The reported results for the small datasets are the accuracy on the test set averaged over the ten splits.
For two large datasets, Yelp and Reddit, we only report the ROC-AUC and accuracy using the publicly available
single set of train, validation, and test sets.
The configuration of the hyper-parameters of~\model~ and additional details for this experiment are presented in the appendix.

The results are reported in Table~\ref{tbl:node:classification:results:on:small:datasets}.
~\model~based on feature and topology augmentations is better than the baselines in almost all datasets.
In addition, it is also comparable with the semi-supervised methods. To highlight the scalability of \model, we evaluate it on large-scale datasets and report the results in Table \ref{tbl:node:classification:results:on:large:datasets}. The self-supervised baselines throw an out-of-GPU memory error for the large datasets. So we  include semi-supervised methods for reference, and not comparison. 

\subsection{Node Classification on Heterogeneous Graphs}
Although \model~is primarily optimized for homogeneous graphs, it can easily be applied to heterogeneous graphs.

Recent studies \cite{li2023multi, ren2019heterogeneous, park2020unsupervised, wang2022collaborative, jing2021hdmi, zhu2022structure, wang2021self} have generalized graph contrastive learning (GCL) to heterogeneous graphs. In these approaches, composite sequences of edge types (i.e., meta-paths) are hand-crafted for an underlying graph to express different possible semantics. For example, a meta-path "author-paper-author" refers to collaboration in a citation graph. Next, meta-path-based graph augmentations are designed for GCL. In doing so, the qualities of the learned representations and augmentation rely heavily on the chosen meta-paths that are typically domain-specific. \model~differs from this line of research, as no pre-defined data augmentation or domain knowledge are assumed.

As a demonstration, we choose a popular dataset commonly used to benchmark methods for heterogeneous GRL.
This dataset, IMDB~\cite{Fu_2020}, has three node types, which are \texttt{movie}, \texttt{director}, and \texttt{actor}, 
and there are two undirected edge types, which are \texttt{movie-to-director} and \texttt{movie-to-actor}.
There are 11,616 ($4,278$--\texttt{movie}, $2081$--\texttt{director}, $5,257$--\texttt{actor}) nodes and 17,106 
($4,278$--\texttt{movie-to-director} and 12,828--\texttt{movie-to-actor}) edges.
The task is to classify the movie nodes as one of the three classes (\emph{Action}, \emph{Comedy}, and \emph{Drama}).

The only modification we need is, instead of a single parameter $\Theta$, we use $\Theta = \lbrace \theta_R \rbrace$, where $\theta_R$ denote the model parameter specific to an edge type $R$.
We use the same experimental setting and splits provided in~\cite{Fu_2020}.
That is, the movie nodes are split into training (400--9.35\%), validation(400--9.35\%), and testing (3,478--81.30\%) nodes.
For the linear evaluation, we only use the test set just as in~\cite{Fu_2020} with different training rates, which are 20\%, 
40\%, 60\%, and 80\%.
For example, when using 20\% for training, we will use 20\% of the test set for training the linear classifier and 
the remainder (80\%) for evaluating and reporting the performance of the learned representations.
In Table~\ref{tbl:node:classification:heterogeneous}, we report the F1-Score of our model and previous methods.
We take the figures for the baselines from~\cite{Fu_2020}, and we see that ~\model~achieves better performance than unsupervised methods and is sometimes comparable to the semi-supervised ones.
Although the paper's primary focus is not on heterogeneous graphs, this experiment highlights the potential of successfully applying a similar approach to this kind of graph.
In future work, we shall address this with more experiments, including more baselines and datasets, and introduce a self-supervised learning technique that generalizes 
not only to homogeneous but also heterogeneous graphs, including knowledge graphs.

\subsection{Graph Property Prediction}

\label{sub:sec:graph:property:prediction:experiment}


In this experiment, we closely follow the experimental protocol suggested for a fair comparison of GNNs in GPP~\cite{errica2020fair}.
Since they provide public splits~\footnote{\url{https://github.com/diningphil/gnn-comparison/tree/master/data_splits}}, we use their split in our experiment.
For each dataset, they provide ten splits, and each split contains a model selection and test splits.
The model selection has training and validation splits.
Similar to the NC experiment, we use the linear evaluation protocol and a similar model selection procedure.
As the number of features for the datasets in this experiment is usually tiny, we also tune $D$ for the baselines between 32 and 128 and for our method between 32 and 64.
Since the social dataset does not have features, we use the degree profile as features.
The configuration of the hyper-parameters of~\model~ and additional details for this experiment are also in the appendix.

The results are reported in Table~\ref{tbl:graph:property:prediction:results}, and we use the published results for semi-supervised methods.
As shown in the table, ~\model~with feature augmentation is better than the baselines in almost all cases.
Although the topology augmentation is comparable with the feature one for the social datasets, it could perform better for
chemical datasets.
We have similar observations for the baselines, which also alter the topology.
Note that even the untrained variants of~\model~are strong competitors for these datasets.
Corroborating the observation in~\cite{trivedi2021augmentations}, that is, what is lost in data augmentation is compensated by 
the strong inductive bias of GNNs.
We believe that corrupting the topology of such datasets requires careful consideration.

\section{Related Work}
\label{sec:related_work}
In general, there have been many frameworks for contrastive learning. 
Mostly, they differ in their data augmentation techniques and the architectures they choose to prevent collapse.

\textbf{Data Augmentation} Although there are well-established data augmentation techniques in the computer vision domain, this is not the case for the graph domain~\cite{DBLP:journals/corr/abs-2006-05582,DBLP:journals/corr/abs-2106-07594}.
Different heuristics, based on high-order networks, perturbation of topology and attributes have been proposed~\cite{velickovic2018deep,DBLP:journals/corr/abs-2006-05582,DBLP:journals/corr/abs-2010-13902,DBLP:journals/corr/abs-2103-14958,bielak2021graph}.
It is unclear what the relative benefit of these augmentation strategies is, and little is known regarding the relevance of each strategy concerning different downstream tasks. Recently, studies~\cite{DBLP:journals/corr/abs-2106-07594,suresh2021adversarial,trivedi2021augmentations} have proposed learnable and contextual augmentation techniques~\cite{suresh2021adversarial,10.1145/3442381.3449802,trivedi2021augmentations}.
However, these methods are restrictive because they either specify a set of graph data augmentation techniques so that the learning 
is choosing the correct technique, or they only learn to dropout edges through adversarial training. A more relevant study, i.e., SimGRACE for graph property classification, \cite{xia2022simgrace}, has perturbed the model weights using Gaussian noise rather than perturbing the node features or the topology. However, a study \cite{li2023multi} has shown that data augmentation in the graph space is more complicated than Gaussian distribution can capture. Therefore SimGRACE learns sub-optimal representations compared to the previously-mentioned data augmentation counterparts. 

On the other hand, recent studies \cite{li2023multi, ren2019heterogeneous, park2020unsupervised, wang2022collaborative, jing2021hdmi, zhu2022structure, wang2021self} have extended graph contrastive learning to heterogeneous networks. However, all these approaches have employed meta-paths to design graph augmentations. Therefore, their performance and the augmentation's quality itself are heavily conditioned on the quality of the manually-chosen meta-paths.

Our study differs from these lines of research, as no predefined data augmentations exist. Secondly, we propose a flexible framework to jointly learn topological or feature augmentation suitable for a given graph.

\textbf{Architectures}
The key difference between existing contrastive architectures arises from the need to prevent trivial solutions.
To this end, existing studies often rely on negative sampling or contrastive terms ~\cite{velickovic2018deep,DBLP:journals/corr/abs-2006-05582,DBLP:journals/corr/abs-2010-13902,10.1145/3442381.3449802,DBLP:journals/corr/abs-2106-07594,sun2021mocl,wang2021selfsupervised,zhu2020deep}.
However, as sampling truly contrastive terms are difficult, other studies have used asymmetric architectures to prevent trivial solutions.
Initially proposed for CV~\cite{DBLP:journals/corr/abs-2006-07733,DBLP:journals/corr/abs-2011-10566}, such methods~\cite{thakoor2021bootstrapped,DBLP:journals/corr/abs-2104-14294,DBLP:journals/corr/abs-2103-14958} have empirically shown that asymmetric networks and a stop gradient operation are sufficient to prevent collapse.
Although the asymmetric methods avoid explicit negative sampling, they are mainly engineering tricks.

Recent studies~\cite{DBLP:journals/corr/abs-2103-03230,bielak2021graph,bardes2021vicreg} have introduced principled approaches based on regularization.
Compared to contrastive architectures with negative sampling, these methods used a principled approach to prevent collapse.
However, in contrast, they do not require explicit negative sampling.

\vspace{-0.2cm}
\section{Conclusion and Discussion}
\label{sec:conclusion}
This paper presents a novel data-driven self-supervised graph representation learning method called~\model. 
Unlike existing methods,~\model~learns augmentation governed by the graph's inherent signal.
We propose two complementary approaches, one based on learning high-order topology and another on learning feature augmentations.
In both cases, augmentation is jointly learned with the graph representation.

We perform an extensive empirical evaluation using eight graph property predictions and another nine node classification datasets, including heterogeneous and homogeneous graphs, which are publicly available.
We compare~\model~against seven popular and SOTA baselines, three for graph property prediction and four for node
classification experiments.
Furthermore, in both experiments, we closely follow recommended protocols for a fair comparison and tuned
the hyper-parameters of all the baselines.
The overall results confirm that DSGRL surpasses the baseline SOTA approaches.

Among the graph property prediction datasets, 4 of them are chemical, and 4 are social datasets.
For the social graphs, the empirical results show that both augmentation techniques produce comparable results.
Whereas for the chemical graphs, the topological augmentation does not perform well.
The latter is also the case for the baselines, which rely on perturbing the topology.
This aligns with existing studies that argue against topological perturbation for such datasets~\cite{trivedi2021augmentations}.
We believe topological augmentations for chemical datasets require further careful investigation.


Last, we report on the untrained variants of the~\model.
Our results show that even the untrained model is significantly better for the chemical datasets than some of
the baselines.
The latter is consistent with recent findings~\cite{trivedi2021augmentations}, which show that the strong inductive bias of GNNs tends to compensate for what is lost in the augmentation.

\bibliography{ecai}

\end{document}